\pdfoutput=1

\documentclass[11pt]{article}

\usepackage[]{acl}

\usepackage{times}
\usepackage{latexsym}
\usepackage{xcolor}

\usepackage[T1]{fontenc}

\usepackage[utf8]{inputenc}
\usepackage{hyperref}

\usepackage{microtype}
\usepackage{booktabs}
\usepackage{graphics}
\usepackage{graphicx}
\usepackage{arydshln}
\usepackage{hyperref}
\usepackage{multirow}
\usepackage{soul}

\newcommand\chbrk{\textsc{ChapterBreak}}
\newcommand\chbrkpg{$\textsc{ChapterBreak}_{PG19}$}
\newcommand\chbrkao{$\textsc{ChapterBreak}_{AO3}$}

\title{\chbrk: A Challenge Dataset for \\ Long-Range Language Models}

\author{Simeng Sun \hspace{3mm} Katherine Thai \hspace{3mm} Mohit Iyyer \\
University of Massachusetts Amherst \\
 \texttt{\{simengsun,kbthai,miyyer\}@cs.umass.edu} \\
}
\begin{document}
\maketitle

\begin{abstract}

While numerous architectures for
 \emph{long-range language models} (LRLMs) have recently been proposed, a meaningful evaluation of their discourse-level language understanding capabilities has not yet followed. To this end, 
we introduce \chbrk, a challenge dataset that provides an LRLM with a long segment from a narrative that ends at a \emph{chapter boundary} and asks it to distinguish the beginning of the ground-truth next chapter from a set of negative segments sampled from the same narrative. A fine-grained human annotation reveals that our dataset contains many complex types of chapter transitions (e.g., parallel narratives, cliffhanger endings) that  require processing global context to comprehend.
Experiments on \chbrk\ show that existing LRLMs fail to effectively leverage long-range context, substantially underperforming a segment-level model trained directly for this task. We publicly release our \chbrk\ dataset to spur more principled future research into LRLMs.\footnote{We make our code and data public at \url{https://github.com/SimengSun/ChapterBreak}}


\end{abstract}
\section{Introduction}

Research on \emph{long-range language models} (LRLMs) aims to process extremely long input sequences by making the base Transformer architecture more efficient (e.g., through sparse attention, recurrence, or cached memory). These modifications are commonly validated by training LRLMs on PG-19~\citep{Rae2020Compressive}, a long-document language modeling dataset, and demonstrating small perplexity decreases over shorter context models~\cite{roy-etal-2021-efficient,Wu2022MemorizingT}.
However, recent analysis experiments~\citep{sun-etal-2021-long,press-etal-2021-shortformer} show that modern LRLMs rely mostly on local context (i.e., the immediately preceding 1-2K tokens) and are insensitive to various perturbations applied to more distant context.

In this paper, we move beyond token-level perplexity by evaluating LRLMs on a task that requires a rich understanding of long-range dependencies. Our task is an instance of \emph{suffix identification}, in which a language model is given a long input sequence (or \emph{prefix}) and asked to disambiguate the next $n$-token segment from a set of hard negatives sampled from the same narrative. To succeed at this task, an LRLM should assign high probability to the ground-truth next segment and low probability to the negatives. To specifically test long-range dependencies, we restrict our prefixes to end at \emph{chapter breaks} of a longer cohesive narrative (e.g., a novel).

\definecolor{prefix}{RGB}{254,250,203}
\definecolor{gsuffix}{RGB}{209,254,211}
\definecolor{nsuffix}{RGB}{250,219,216}
\setlength{\fboxsep}{0pt}
\begin{figure}[t]
    \centering
    \includegraphics[width=0.99\linewidth]{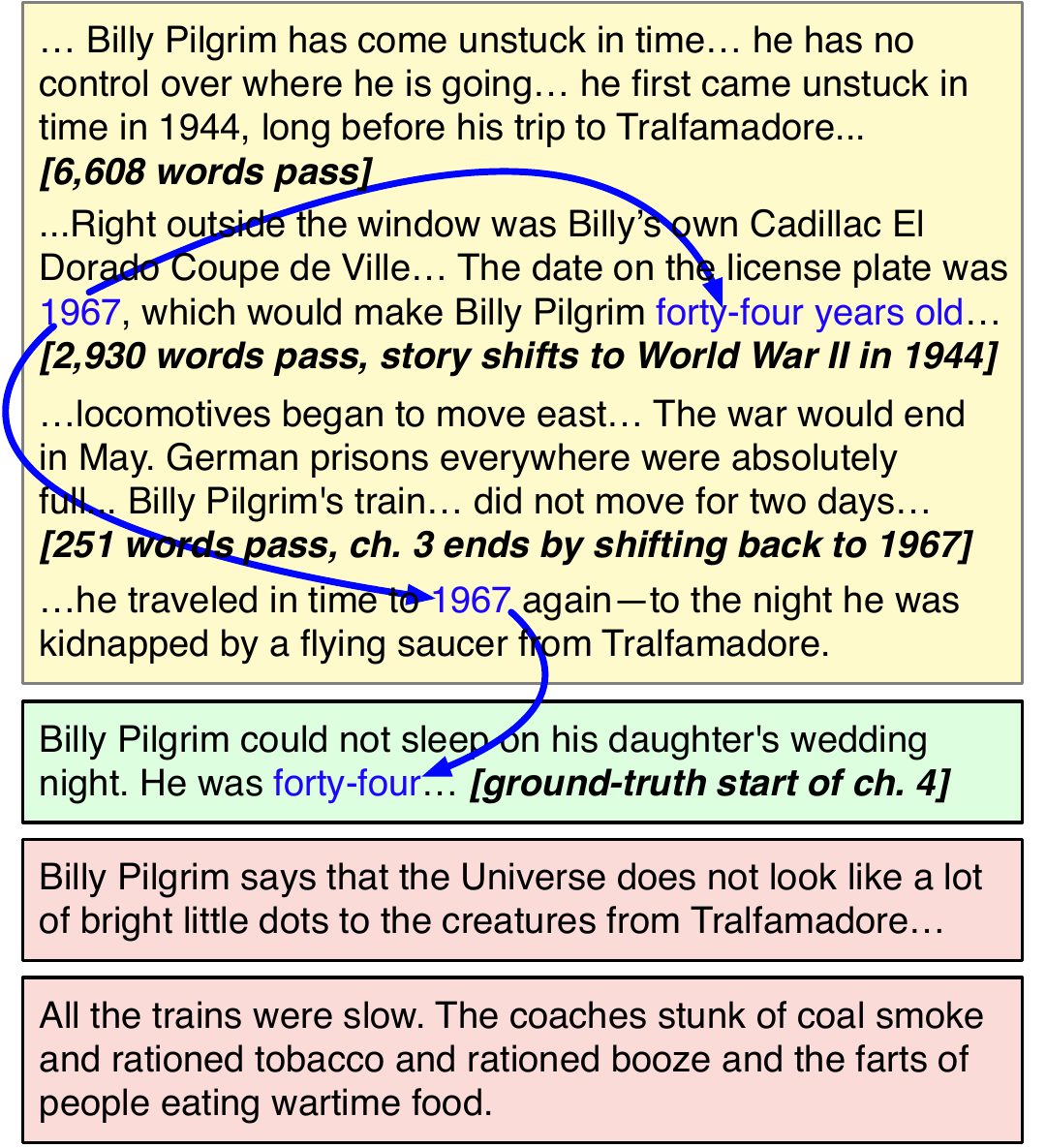}
    \caption{An illustrative example of our suffix identification task from Kurt Vonnegut's \emph{Slaughterhouse-Five}, in which an LRLM needs to make connective inferences across temporal and spatial shifts in a long \colorbox{prefix}{\strut prefix} of the narrative to correctly disambiguate the \colorbox{gsuffix}{\strut start of the next chapter} from \colorbox{nsuffix}{\strut negative examples}.}
    \label{fig:slaughter}
\end{figure}

We construct a challenge dataset, \chbrk, by automatically detecting chapter boundaries within both held-out PG-19 documents (in-domain for pretrained LRLMs) and works of fan fiction published on the Archive of Our Own (out of domain).\footnote{\url{{https://archiveofourown.org}}} We perform a detailed analysis of the types of chapter transitions in our dataset and discover a high frequency of narrative shifts in point-of-view, location, and time, all of which require global narrative understanding over long input sequences. For example, Figure~\ref{fig:slaughter} contains a complex prefix in which the time-traveling Billy Pilgrim moves between World War II, 1960s suburban life, and an alien planet. Understanding the cliffhanger ending, in which the narrative abruptly switches from a wartime scene to a 1967 alien abduction, requires an LRLM to make connective inferences using details buried far back in the context (e.g., Billy's age in 1967).

We evaluate three LRLMs on \chbrk, including BigBird~\citep{zaheer2020big}, the Routing Transformer~\citep{roy-etal-2021-efficient}, and its local attention variant, all pretrained or fine-tuned on PG-19. Our experiments show that these LRLMs perform poorly at selecting the ground-truth suffix, regardless of the length of the input sequence. 
As an upper bound, we train a small RoBERTa-based segment-level language model on PG-19 and discover that it substantially outperforms all LRLMs on \chbrk, which suggests that LRLMs have considerable room for improvement on this suffix identification task. Finally, we perform an analysis on the instances for which all models struggle to choose the correct suffix, which shows that  shifts in location and events in focus are particularly challenging to disambiguate. Taken together, these results suggest that \chbrk\ is a useful benchmark for future research into LRLMs. 

\section{The \chbrk\ dataset} \label{sec:dataset}

Authors often break long-form narratives into a sequence of discrete chapters to impose ``an order and shape over events in time''~\citep{stevick1970chapter}. 
Henry Fielding writes in his novel \emph{Joseph Andrews} that the space between chapters is like ``an Inn or Resting Place'' for readers to reflect on the preceding chapter~\citep{fielding1779history}. 
Chapters come in many flavors:  for example, Murakami's \emph{Kafka on the Shore} uses chapter breaks to alternate between parallel narratives focusing on the two protagonists, while cliffhanger endings such as the one in Figure~\ref{fig:slaughter} add suspense. Making sense of the complex narrative shifts associated with chapter transitions (e.g., changes in point-of-view, time, location, and theme) requires a deep understanding of the entire text.  To maintain global narrative coherence, ~\citet{myers-global} show that human readers tend to reactivate memory about ``backgrounded'' information from the long-range context.

\paragraph{Task overview:}
Given that chapter transitions requires global context understanding, how can we turn this into a task to evaluate LRLMs? A simple approach is to evaluate the token-level perplexity of an LRLM only at chapter boundaries (i.e., on the first $n$ tokens of each chapter); however, the vast majority of tokens can be predicted using just local context~\citep{sun-etal-2021-long} under the teacher-forcing setup, which obscures an LRLM's usage of long-range context as we show in Section~\ref{sec:experiments}. We instead turn to the task of \emph{suffix identification}, which closely resembles existing datasets such as SWAG~\citep{zellers-etal-2018-swag}. 

Each instance of our task is defined by a triplet $(c, s^+, s^-_i \in \textbf{N})$, where $c$ is a prefix sequence of up to 8K tokens that ends at a chapter break, $s^+$ is the gold suffix of length 128 tokens (i.e., the beginning of the next chapter), and $s^-_i$ is a negative 128-token-long suffix from a set $\textbf{N}$ of five\footnote{We use a small number of negatives because it is time-consuming and resource-intensive to evaluate the probabilities of long sequences with LRLMs.} future chapter beginnings sampled from the same narrative.\footnote{In Appendix~\ref{sec:in-vs-out}, we show that in-book negatives are much harder than out-of-book negatives as they often contain the same named entities and rare tokens as the gold suffix. Thus, disambiguating the correct suffix requires a deep understanding of the context.}  All negatives are modified to begin with the same chapter index (e.g., if the gold suffix begins with ``Chapter III'', the chapter indices of all negatives is set to ``Chapter III'') to eliminate the effect found by~\citet{sun-etal-2021-long} of language models memorizing chapter indices in long contexts.
We then evaluate whether an LRLM assigns higher probability to the gold suffix $P(s^+|c)$ than to all negative suffixes $P(s^-_i|c)$.

\begin{table}[]
    \centering
    \footnotesize
    \scalebox{0.88}{
    \begin{tabular}{@{}p{0.15\linewidth}p{0.75\linewidth}p{0.1\linewidth}@{}}
\toprule
                      \textbf{Category}       &            \textbf{Definition}                                            & \textbf{Pct.} \\ \midrule
\multirow{3}{*}{Events}      & Previous event ends and new event starts               & 76\%             \\
                             & Previous event continues into next chapter & 24\%             \\ \midrule
\multirow{2}{*}{Actors}      & Change of perspective or character in focus            & 36\%             \\
                             & No change in POV or main character                     & 64\%             \\ \midrule
\multirow{2}{*}{Locations}   & Change of location                                     & 68\%             \\
                             & No change in location                                  & 32\%             \\ \midrule
\multirow{5}{*}{Continuity} & Discontinuous but chronological                        & 29\%             \\
                             & Continuous                                        & 62\%             \\
                             & Analepsis                                              & 2\%              \\
                             & Parallel                                          & 6\%              \\ \bottomrule
\end{tabular}
    }
    \caption{Our human annotation on 300 chapter transitions randomly sampled from \chbrkao\ shows the diversity and complexity of the dataset.}
    \label{tab:annot_stats}
\end{table}

\paragraph{Dataset overview:} Where do we get these triplets from? We collect a dataset, \chbrk, with two splits: \chbrkpg, which contains $241$ examples extracted from the PG-19 validation set~\cite{Rae2020Compressive},\footnote{We only collect examples from validation set as two baseline models in the later sections are trained on PG-19.} and \chbrkao, which contains 7,355 examples extracted from an online dump\footnote{\url{https://archive.org/download/AO3_story_dump_continuing}} of fanfiction posted on Archive of Our Own (AO3). We apply filtering to remove fanfiction works that are too short or not rated for general audiences.
Each work contains on average 42K words and 21.5 chapters.\footnote{More preprocessing details and statistics can be found in Appendix~\ref{sec:data-stats}.} Even though the \chbrkpg\ split is small, we include it because many LRLMs are pretrained on PG-19; the much larger \chbrkao\ split is out-of-distribution for all models that we evaluate. To extract chapters in PG-19, we match for lines beginning with the string ``chapter'', while AO3 stories already have chapter-level metadata. 


 \paragraph{What are the different types of transitions in \chbrk\ and how often do they occur?} To get a better sense of our dataset, we perform a fine-grained annotation of 300 randomly-selected chapter transitions from \chbrkao. For each transition, we annotate any changes in the following four aspects: events, actors (characters in focus), locations, and continuity. To annotate continuity, we follow a simplified version of the scheme proposed by~\citet{ireland1986towards},\footnote{To validate our continuity annotations, we also annotate every chapter in \textit{Pride and Prejudice} and obtain almost the same proportion of continuous transitions (67\%) as the number reported by the expert annotation of~\citet{ireland1986towards} (72\%).} which considers five categories: \textbf{continuous} (the next chapter occurs within a day of the previous chapter), \textbf{discontinuous} (the next chapter occurs more than a day after the previous chapter), \textbf{analepsis} (the next chapter is a ``flashback'' to an earlier point in the narrative), 
 and \textbf{parallel} (the next chapter reverts to the time of a previous chapter, switching the character or event in focus).\footnote{ Appendix~\ref{sec:annotation} contains more details about each category.} The results, shown in Table~\ref{tab:annot_stats}, demonstrate that \chbrk\ covers a diverse array of transitions, including many that require global narrative understanding.


\begin{table}[t]
    \centering
    \footnotesize
    \scalebox{0.83}{
    \begin{tabular}{lccccc}
    \toprule
         & \#Params & Seq Len & PPL$_{\textnormal{\scriptsize PG19}}$ & Acc$_{\textnormal{\scriptsize PG19}}$ & Acc$_{\textnormal{\scriptsize AO3}}$ \\\midrule
        LT & 516M & 8K &    76.8   & 25\%  & 24\% \\
        RT & 490M &   8K &   72.3     & 22\%  & 24\% \\
        Bigbird & 128M &  4K &   56.2     & 27\%  & 26\% \\
        \midrule
        GPT-2 & 1.5B & 1K &   78.2     & 23\%  & 24\% \\
        GPT-3 & 175B &  2K &    -    & 36\%$^*$ & 28\%$^*$\\
        \midrule
        SuffixLM & 87M &  10K &    -    &\textbf{52\%} & \textbf{41\%}\\
        \bottomrule
    \end{tabular}
    }
    \caption{Summary of LRLMs (top), Transformer LMs (middle), and our SuffixLM (bottom). All models are trained or fine-tuned on PG-19 except for GPT-2. The third column shows the word-level perplexity of gold suffix in the PG-19 split. The last two columns show the suffix identification accuracy of each model on the two \chbrk\ splits when evaluated at maximum input length. $^*$ indicates results are on a subset of \chbrk.}
    \label{tab:model_summ}
\end{table}

\begin{figure}[t]
    \centering
    \includegraphics[width=\linewidth]{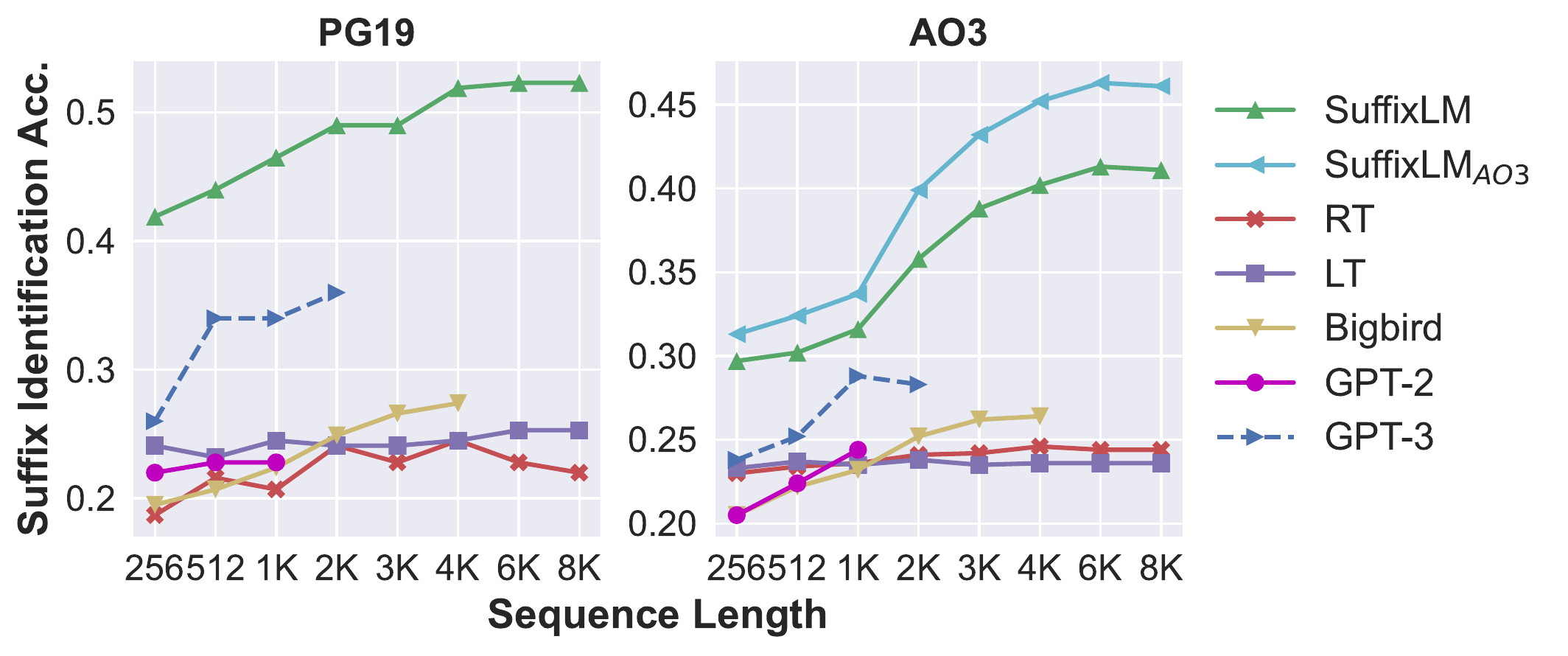}
    \caption{Suffix identification accuracy on both splits (PG-19 and AO3) of \chbrk\ is much lower for LRLMs than our SuffixLM upper bound.}
    \label{fig:chbrk_only}
\end{figure}

\section{Experiments} \label{sec:experiments}
We evaluate three different long-range language models on \chbrk\ and compare their results to those of standard Transformer language models as well as an upper bound directly trained for suffix prediction.

\paragraph{Language models:} We evaluate three LRLMs pretrained on PG-19: the Local Transformer~\citep[][LT]{roy-etal-2021-efficient}, Routing Transformer (RT)~\citep[][RT]{roy-etal-2021-efficient}, and BigBird~\citep{zaheer2020big}. The BigBird model is the decoder part of the released checkpoint fine-tuned with causal LM objective on 14k books of PG-19 for 100k steps. We also evaluate two standard Transformer language models, GPT-2 large~\cite{radford2019language} and GPT-3~\cite{brown2020language}.\footnote{Due to OpenAI's API costs for GPT-3, we only evaluate in total a subset of 200 examples instead of the full dataset.} We summarize these models in Table~\ref{tab:model_summ}, more details about each model are included in Appendix~\ref{sec:background}.

\paragraph{An upper bound directly trained for suffix identification:}
As authors often write stories that are intended to surprise readers, it is possible that many examples in \chbrk\ are ambiguous by nature (i.e., the upper bound for suffix identification accuracy may not be 100\%). To obtain a reasonable upper bound, we also train a model (SuffixLM) directly on the suffix identification task by scaling up the sentence-level language model proposed by~\citet{ippolito-etal-2020-toward}.\footnote{Our SuffixLM can process up to 10K tokens, while the model of~\citet{ippolito-etal-2020-toward} supports only up to ten sentences. } We divide an input sequence into multiple segments, each of which is embedded via the \texttt{[CLS]} vector of a small fine-tuned RoBERTa network~\citep{liu2019roberta}.  Our SuffixLM then performs ``language modeling'' atop the dense \texttt{[CLS]} vectors, predicting the next segment representation given the representations of previous segments via contrastive predictive coding~\cite{Oord2018RepresentationLW}.\footnote{Our SuffixLM is closely related to the model in~\citet{ainslie2020etc}, but differs crucially by predicting the representation of next segment instead of summaries.} Formally, our SuffixLM minimizes the following loss:
\[\mathcal{L}_i = -\log\frac{\exp(\hat{\mathbf{z}_{i}}^\top \mathbf{z}_i^+)}{\sum_{\mathbf{z}_i \in \{\mathbf{z}_i^+, \mathcal{Z}_{i}^-\}}\exp(\hat{\mathbf{z}_{i}}^\top \mathbf{z}_i)}\]
where $\hat{\mathbf{z}_{i}}$ is the predicted representation by SuffixLM, $\mathbf{z}_{i}^+$ is the gold suffix representation obtained from a small encoder (RoBERTa), and $\mathcal{Z}_{i}^-$ is the set of dense representations of the negatives. More details about our SuffixLM are included in Appendix~\ref{sec:best-SuffixLM}.

\begin{figure}[t]
    \centering
    \includegraphics[width=0.49\linewidth]{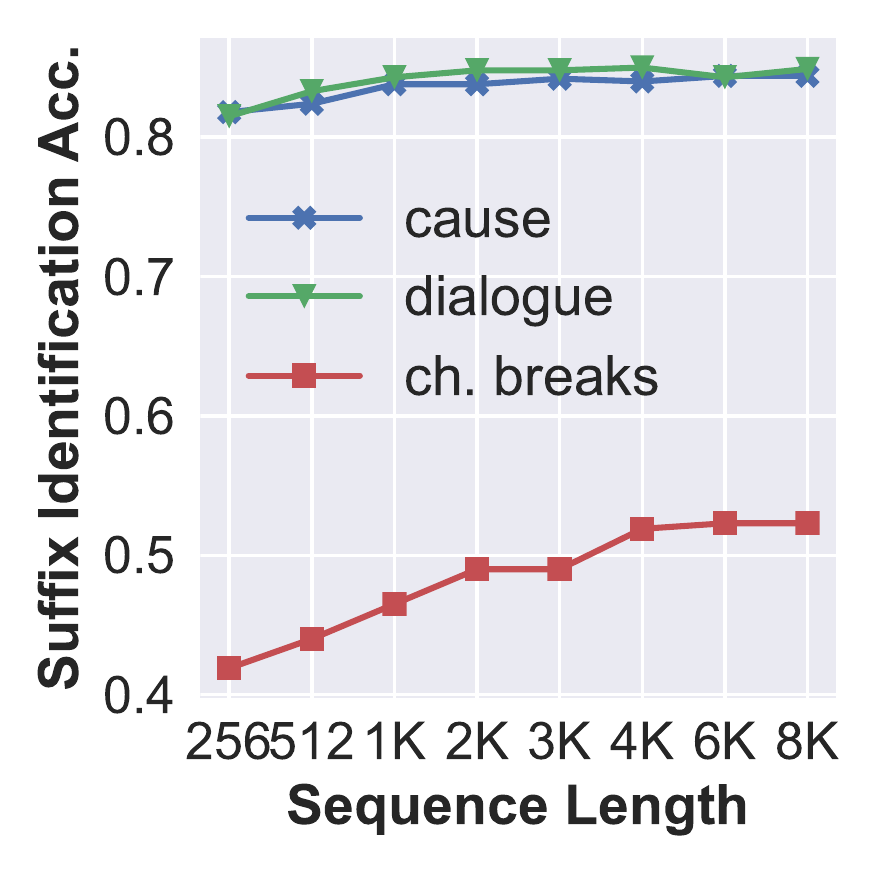}
    \includegraphics[width=0.49\linewidth]{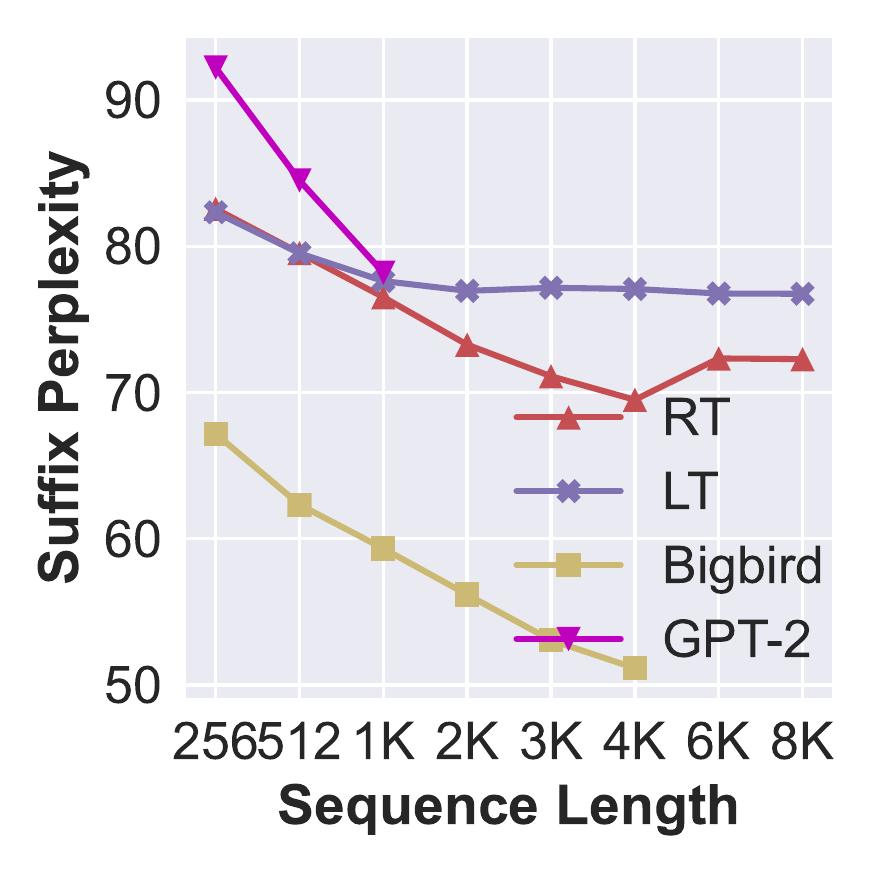}
    \caption{\textbf{Left:} Prefixes ending at chapter breaks benefit more from long-range context than other types of discourse boundaries. \textbf{Right:} Word-level perplexity of the gold suffix does not correlate to accuracy (e.g., GPT-2 has high perplexity but outperforms RT on suffix identification).}
    \label{fig:rank_analysis}
\end{figure}

\section{Results \& Analysis}
Overall, the results in Table~\ref{tab:model_summ} (rightmost two columns) confirm that all of the language models studied in this paper struggle on \chbrk, especially when compared to the SuffixLM upper bound, which outperforms the best LM by $\sim$25\% absolute accuracy when evaluated on the entire PG-19 split. We describe other interesting results and analysis below:

\paragraph{Accuracy increases with longer prefixes:} Figure~\ref{fig:chbrk_only} shows that as prefix sequence length increases, some LRLMs (e.g., LT) barely improve, while others show modest improvements (e.g., GPT-3 and fine-tuned BigBird). However, all LRLMs significantly underperform our SuffixLM upper bound, even when the SuffixLM is given prefixes that are only 256 tokens long. Additionally, SuffixLM's accuracy increases far more than those of LRLMs when increasing the prefix length (from 31\% at prefix length of 256 to 46\% at 8K on the AO3 split\footnote{We collected 13,682 fan-fictions posted on AO3 and fine-tuned our SuffixLM on subset of this dataset to be the model SuffixLM$_{AO3}$. More details about the filtered AO3 works are included in Appendix~\ref{sec:data-stats}}). This result suggests that the token-level LRLMs evaluated in our work are not taking full advantage of information in the long-range context to solve \chbrk.

\paragraph{Perplexity does not always correlate with accuracy:} Previous LRLM efforts use validation perplexity (e.g., on PG-19) to compare against other models. However, we show that perplexity is not by itself a predictor of suffix identification accuracy: As shown in Table~\ref{tab:model_summ}, GPT-2 achieves higher accuracy than RT despite yielding a word-level perplexity of 78.2 on gold suffixes, compared to 72.3 for RT.\footnote{As these models use different tokenizers, we normalize the subword-level perplexities to the word level as suggested by~\citet{Rae2020Compressive}. More details about this can be found in Appendix~\ref{sec:suffix-ppl}.} We advocate that future research on LRLMs includes evaluation on suffix identification tasks like \chbrk, as perplexity alone does not reflect LRLMs' capabilities to model long-range dependencies. 


\paragraph{Why chapter breaks over other discourse boundaries?} Other discourse markers, including \emph{cause} and \emph{dialogue}, also often prompt human readers to reactivate memories of global context~\citep{Albrecht1995RoleOC}. We create suffix identification datasets for these two discourse markers by string matching over corresponding cue phrases (`because', `due to' for the \emph{cause} subset and text within quotation marks for \emph{dialogue}).\footnote{Appendix~\ref{sec:data-stats} contains more details about data for these two discourse markers.} Figure~\ref{fig:rank_analysis} (left) shows that with prefixes of length 256 tokens, our SuffixLM is able to successfully disambiguate the correct suffixes for both discourse markers more than 80\% of the time, while the accuracy is much lower at chapter boundaries. As the prefix length increases, accuracy only slightly increases for \emph{cause} and \emph{dialogue}, especially compared to the robust improvement at chapter boundaries.\footnote{ Appendix~\ref{sec:multi-suffix-type} shows similar trends on \emph{cause} and \emph{dialogue} with other models.}


\paragraph{Short-context Transformers are comparable to LRLMs:}
Our results show that GPT-2, despite its high perplexity on gold suffixes and short maximum sequence length (1024 tokens), achieves comparable performance to RT and LT on both splits. Meanwhile, GPT-3 achieves much higher performance on both \chbrk\ at a sequence length of 2,048 tokens, and the increasing GPT-3 curve in Figure~\ref{fig:chbrk_only} is promising for future work scaling LMs to longer sequence lengths.

\paragraph{Limitations of our work:} While we have used the SuffixLM as an upper bound in this paper and demonstrated that it substantially outperforms LRLMs on \chbrk, a more compelling comparison would include human performance on our task at varying prefix lengths, especially since some chapter transitions are specifically intended by their authors to be unpredictable. However, obtaining reliable human performance numbers is very difficult, as it requires in-depth comprehension of long narratives on the part of workers. Due to the time-consuming nature of this task and its high cognitive demand, it is not possible (within a reasonable budget) to use crowdsourcing, as ensuring that the annotators fully read the prefix instead of skimming or ignoring it is a major challenge. These issues also carry over to experiments performed with in-person subjects. As such, we leave a thorough human evaluation on \chbrk\ to future work.

\section{Related Work}

Our work depends heavily on recent advances in efficient Transformers~\cite{tay2020efficient} that process long sequences~\cite{ Rae2020Compressive,beltagy2020longformer,zaheer2020big,ainslie2020etc,roy-etal-2021-efficient}. 
Sparse attention~\cite{child2019generating}, relative position encoding~\cite{shaw-etal-2018-self,raffel2020exploring,guo2021longt5}, recurrence mechanism and memory~\cite{dai-etal-2019-transformer,weston2015memory,Hutchins2022BlockRecurrentT,Wu2022MemorizingT} and other tricks~\cite{Shen_Dong_Ye_Ma_Yao_Gholami_Mahoney_Keutzer_2020,katharopoulos20,gupta2020gmat,stock2021training,yogatama-etal-2021-adaptive,borgeaud2021improving,hawthorne2022general} are commonly adopted by recent Transformer variants to make the operation on long sequences more time/memory efficient. 

Besides perplexity, many downstream extrinsic tasks for evaluating long-range language models were developed recently , such as long-form QA~\cite{fan2019eli5,pang2021quality}, document-level summarization~\cite{kryscinski2021booksum,huang-etal-2021-efficient}, and machine translation~\cite{liu-zhang-2020-corpora}.  More recently, ~\citet{shaham2022scrolls} introduce a new benchmark covering multiple domains and tasks, while~\citet{tay2021long} propose multimodal long sequence tasks. 
\section{Conclusion}

We introduce \chbrk, a suffix identification dataset targeted at evaluating the discourse-level understanding of long-range language models. The dataset is extracted from long-form narratives and covers a variety of complex chapter transitions, such as shifts in location and events in focus. Experiments show that existing LRLMs perform poorly on \chbrk\ and much worse than a SuffixLM trained as an upper bound on this task. We release the dataset to spur more principled development of future LRLMs.


\section*{Acknowledgements}

We thank the anonymous reviewers and UMass NLP group for the thoughtful comments on the draft of this paper. We are grateful to AO3 Support Chair and volunteers for answering data related questions. This work was supported by
awards IIS-1955567 and IIS-2046248 from the National Science Foundation (NSF).

\section*{Ethical Considerations}

\chbrk\ is constructed from two sources: public domain books published prior to 1919 (from the held-out set of PG-19) and works of fanfiction extracted from an online dump of stories posted on Archive of Our Own (AO3). We refer readers to ~\citet{Rae2020Compressive} for more details about PG-19. For AO3, we apply multiple filters to obtain long fanfiction stories rated as suitable for ``General Audiences''. We refer readers to Appendix~\ref{sec:data-stats} for more preprocessing details. More generally, this work focuses on long-range language models, which could potentially be misused to generate offensive output. However, the main purpose of this paper is to present a dataset which provides a better evaluation of the discourse-level capabilities of such models.


\bibliography{anthology,custom}
\bibliographystyle{acl_natbib}
\newpage
\appendix
\section{Dataset statistics} \label{sec:data-stats}

We collected 13,682 fanfictions from an online dump of stories posted on Archive of Our Own (AO3) by filtering works written in English language, rated General Audience by the author and contains at least 10K words and more than 10 chapters. For each chapter, we remove the text within the range of ``\texttt{**Notes for the Chapter:**}'', ``\texttt{**Summary for the Chapter:**}'' and ``\texttt{**Author's Note:**}''. The meta-comments inserted into the main text by the authors are not removed. The statistics of this long-fic dataset are included in Table~\ref{tab:ao3_longfic}. We do not apply  other profanity filters to the fictions, therefore there may still be inappropriate content for general audience as the rating is self-labeled by each author. Besides chapter breaks introduced in the main text, we also collected two other discourse boundaries, cause and dialogue, as comparisons to the chapter boundary examples. We present the statistics each type of examples in Table~\ref{tab:eval-data-stats}.
\begin{itemize}
    \item \textbf{Cause}: The beginning of the suffix contains words or phrases `because', `due to', `owing to'. According to~\cite{Albrecht1995RoleOC}, human readers reactivate memory of global context for comprehending statements following causes or goals.
    \item \textbf{Dialogue}: The gold suffix in this category starts with a quotation mark. This often happens in dialogues where the continuation of one interlocutor depends heavily on the immediately preceding utterance. We conjecture this is the type where the prediction relies more on the local rather than the global context. 
\end{itemize}

\begin{table}[!h]
    \centering
    \begin{tabular}{cccc}
    \toprule
      &  mean  & min  &  max   \\ \midrule
    \#chapters & 21.5 & 11  & 589 \\
   \#words  &  41,513.2 & 10,000 & 636,468 \\
      \bottomrule
    \end{tabular}
    \caption{Statistics of long fanfictions collected from AO3 story dump.}
    \label{tab:ao3_longfic}
\end{table}

\begin{table}[!h]
\scalebox{0.8}{
\begin{tabular}{@{}lcccc@{}}
\toprule
               & \multicolumn{2}{c}{AO3} & \multicolumn{2}{c}{PG19} \\ \midrule
Suffix Type    & \#works   & \#examples  & \#works   & \#examples   \\ \midrule
cause          & 965       & 8,133        & 45        & 506          \\
dialogue       & 979       & 8,724        & 46        & 3,165         \\
chapter breaks & 1202       & 7,355        & 17        & 241          \\ \bottomrule
\end{tabular}
}
\caption{Data statistics of \chbrk\ as well as another two discourse boundary examples.}
\label{tab:eval-data-stats}
\end{table}

\section{Annotation Scheme} \label{sec:annotation}

We annotate each chapter transition from four aspects: events, actors (point-of-view or characters in focus), location, and continuity in timeline. 

\paragraph{Events} We define two subcategories based on whether (1) previous event ends in the previous chapter and new event starts in the new chapter, (2) old event does not end and continues into the next chapter.
\paragraph{Actors} We define two subcategories based on whether there is a shift in POV or main character in focus.
\paragraph{Location} We define two subcategories based on whether the location described in the prefix and in the new chapter is different.
\paragraph{Continuity} Following~\citet{ireland1986towards}'s work, we categorize the chapter transition by timeline continuity into four subcategories: 
\begin{itemize}
    \item \textbf{Discontinuous but chronological}: Reusing the standard by~\citet{ireland1986towards}, discontinuous represents a gap in time forward for more than one night.
    \item \textbf{Continuous}: The time interval between chapters lasts for no more than one night.
    \item \textbf{Analepsis}: Analepsis represents retrospective evocation of an event, or ``flashback'' to an earlier point in the narrative.
    \item \textbf{Parallel}: This includes timeline reverting back to the time of any previous chapter, typically accompanied by switching character in focus or description of a separate set of events independent of the last chapter. This category is a collapse of ``alternate phase'', ``parallel phase'' and ``simultaneous phase'' introduced in~\cite{ireland1986towards}.
\end{itemize}

\section{Baselines} \label{sec:background}


\paragraph{Bigbird~\cite{zaheer2020big}} To reduce the quadratic complexity of self-attention in the standard Transformer, the Bigbird model employs a mixture of global, random and local attention mechanisms, which successfully reduce the complexity to linear. The idea is to insert each sequence $O(1)$ global tokens, which attend to all other tokens. The rest tokens attend to their neighbor tokens, random tokens in the sequence as well as the inserted global tokens. A very similar idea is developed concurrently in the Longformer~\cite{beltagy2020longformer}. The Bigbird model we fine-tuned is the decoder part of the released checkpoint. We fine-tune the model with causal LM objective on 14K books of PG-19 with peak learning rate $0.0001$ for 100K steps. We set attention type to be ``original\_full'' instead of using ``block\_sparse'' during fine-tuning. Training is completed on a single RTX8000 GPU for around 6 days.

\paragraph{Local Transformer} Rather than implementing all three types of sparse attention in Bigbird, the Local Transformer relies only on the local attention, i.e., each token attends to neighbors within a local window. The maximum attainable sequence length scales linearly with the number of layers, e.g., with window size $k$, the token representation at layer $l$ theoretically covers information in a range of $k \times l$ tokens. 

\paragraph{Routing Transformer~\cite{roy-etal-2021-efficient}} Different from previously described models which use \emph{position}-based sparse attention, the Routing Transformer employs \emph{content}-based sparse attention. Namely, each token are routed to clusters and the attention is performed only within each cluster. The clustering operation effectively reduces the quadratic complexity in length $L$ to $O(L^{1.5})$. Both the RT and LT checkpoint we used were trained on PG-19~\cite{Rae2020Compressive}. For both RT and LT, we evaluate on single RTX8000 GPU. 

\paragraph{GPT-2/3} The GPT models have a lot shorter maximum input length than the rest models we evaluated. While GPT-2 model does not use sparse attentions at all, GPT-3 model adopts alternated layers of sparse and dense self-attention. We use the GPT-2 large model, which was pre-trained on data scraped from the Internet. The GPT-3 model was pre-trained on a mixture of filtered CommonCrawl, WebText2, Books1, Books2, and Wikipedia.

\begin{figure}[t]
    \centering
    \includegraphics[width=0.5\textwidth]{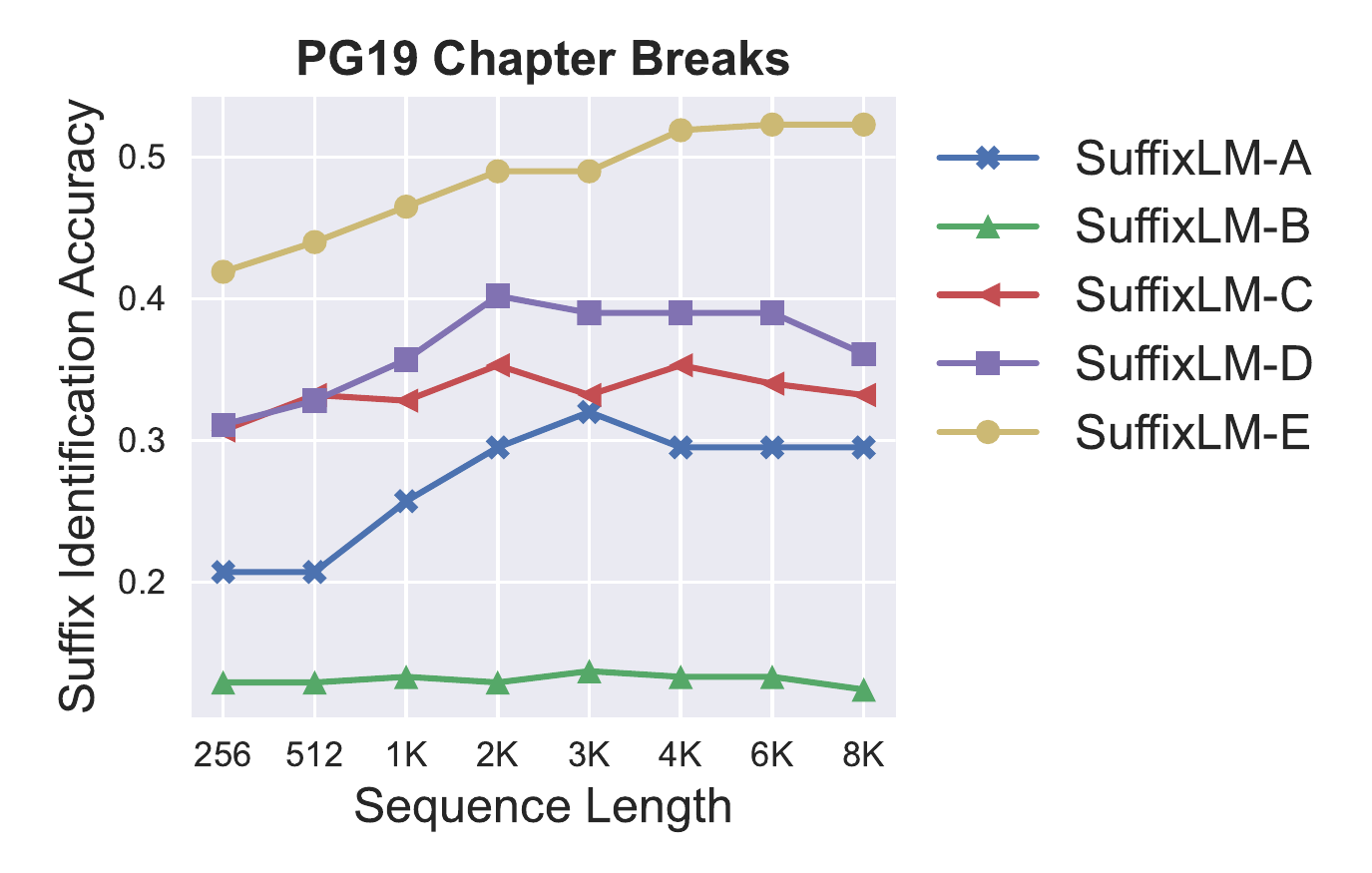}
    \caption{Performance of each SuffixLM variant. Detailed information about each variant is included in Appendix~\ref{sec:best-SuffixLM}.}
    \label{fig:SuffixLM_variants}
\end{figure}

\begin{figure*}
    \centering
    \includegraphics[width=0.8\textwidth]{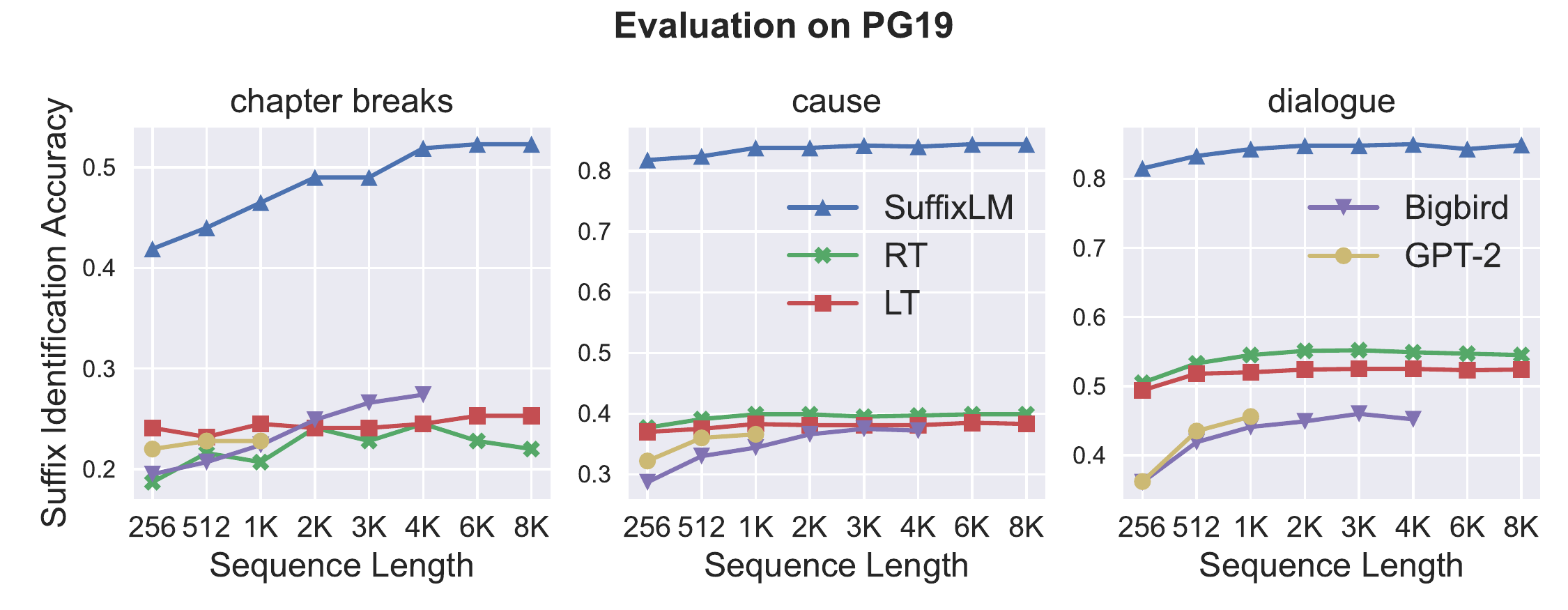}\\
    \includegraphics[width=0.8\textwidth]{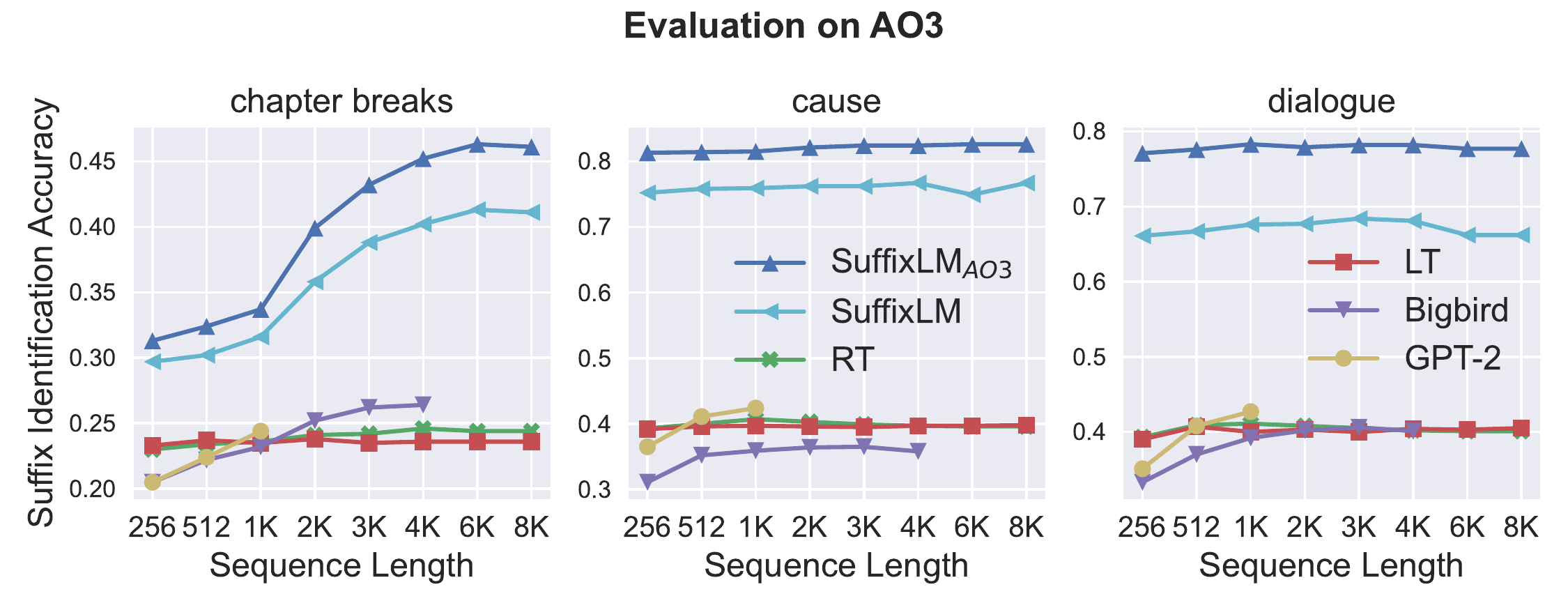}
    \caption{Evaluation results on both \chbrkpg\ and \chbrkao. }
    \label{fig:all_types}
\end{figure*}

\section{Finding the best SuffixLM} \label{sec:best-SuffixLM}

As there are no prior long-range segment-level LM architectures that we can borrow from, we experiment multiple design choices and report the result of only the best performing one in the main text. For all variants, we use RoBERTa-base~\cite{liu2019roberta} as the encoder to obtain the encoded segment representation. This is done by extracting the representation of the $[CLS]$ token prepended at the beginning of each sequence. We describe five variants below.
\begin{itemize}
    \item \textbf{SuffixLM-A} This variant contains a frozen RoBERTa-base encoder and a SuffixLM using a 6-layer Transformer as the base architecture. 
    \item \textbf{SuffixLM-B} This variant contains a frozen RoBERTa-base encoder and a SuffixLM using a 6-layer average-attention Transformer as the backbone. The motivation of using uniform distribution for attention weights is to encourage the model to get more information from the distant context rather than rely too much on local context.
    \item \textbf{SuffixLM-C} This variant is essentically SuffixLM-A but during training we perform ``segdrop'' -- stochastically dropping prefix segments with probability 0.2\footnote{Tried \{0.1, 0.2, 0.4\}, 0.2 works the best.} when performing self-attention. When the local segments are dropped, the model has to predict the next segments with only the distant context, which also encourages learning better long-range prefix representations.
    \item \textbf{SuffixLM-D} Instead of freezing the encoder, this variant fine-tunes part of the encoder and the rest is the same as SuffixLM-A. Due to limited memory capacity, we only fine-tune the last two layers of the RoBERTa-base. 
    \item \textbf{SuffixLM-E} This model is the same as SuffixLM-D except that we truncate the encoder to just the two tunable layers and train all parameters in the encoder including the embedding parameters.
\end{itemize}
All SuffixLMs with frozen encoders are trained with average sequence length of 10240 tokens for up to 60k steps, and the one with trainable encoder is trained for max 120k steps. The dimension of the model is 768, hidden dimension 2048,attention heads 8. The peak learning rate is $0.0001$ with warm up steps 4000. We train SuffixLM on entire PG-19 dataset and evaluate the best checkpoint selected by dev loss. We use segment size 128 in all SuffixLMs we trained. Each segment starts from a new sentence, if not reaching 128 tokens, we pad with a special `<pad>' token. For very long sentences, the part exceeding 128 tokens overflows to the next segment. We plot the suffix identification accuracy of each variant on \chbrk\ while feeding in prefixes of increasing length. As shown in Figure~\ref{fig:SuffixLM_variants}, SuffixLM-E outperforms all other variants across various prefix lengths. Therefore in the main text, all SuffixLM refers to the SuffixLM-E variant. Note that one limitation of SuffixLM is it exclusively models on segment-level, which prohibits it from performing token-by-token generation and thus impossible for us to evaluate perplexity.

\section{Suffix perplexity} \label{sec:suffix-ppl}

Although the task of \chbrk\ is to identify gold suffix from negatives, we also present the gold suffix perplexity of next-token prediction LMs. Note that all models were trained or fine-tuned on PG-19 except for GPT-2/3. As these models use different tokenizers, the $128$-token suffix may cover different number of words, to make the results comparable, we convert the subword-level perplexity to word-level by multiplying a constant to the log probability value of each model. For RT/LT, we multiple by 1.248 as used in the official repository. We multiply the value by 1.30 for GPT-2, and 1.22 for Bigbird. These values are estimated via the subword/word ratio on validation set of PG-19. Our fine-tuned Bigbird model achieves the lowest perplexity on PG-19, even better than Routing Transformer or Local Transformer. This implies that context from long-range is not necessary for achieving low perplexity since the maximum input length of Bigbird is half that of RT/LT. 


\section{In-book vs. Out-of-book}\label{sec:in-vs-out}
\begin{figure}[t]
    \centering
    \includegraphics[width=0.245\textwidth]{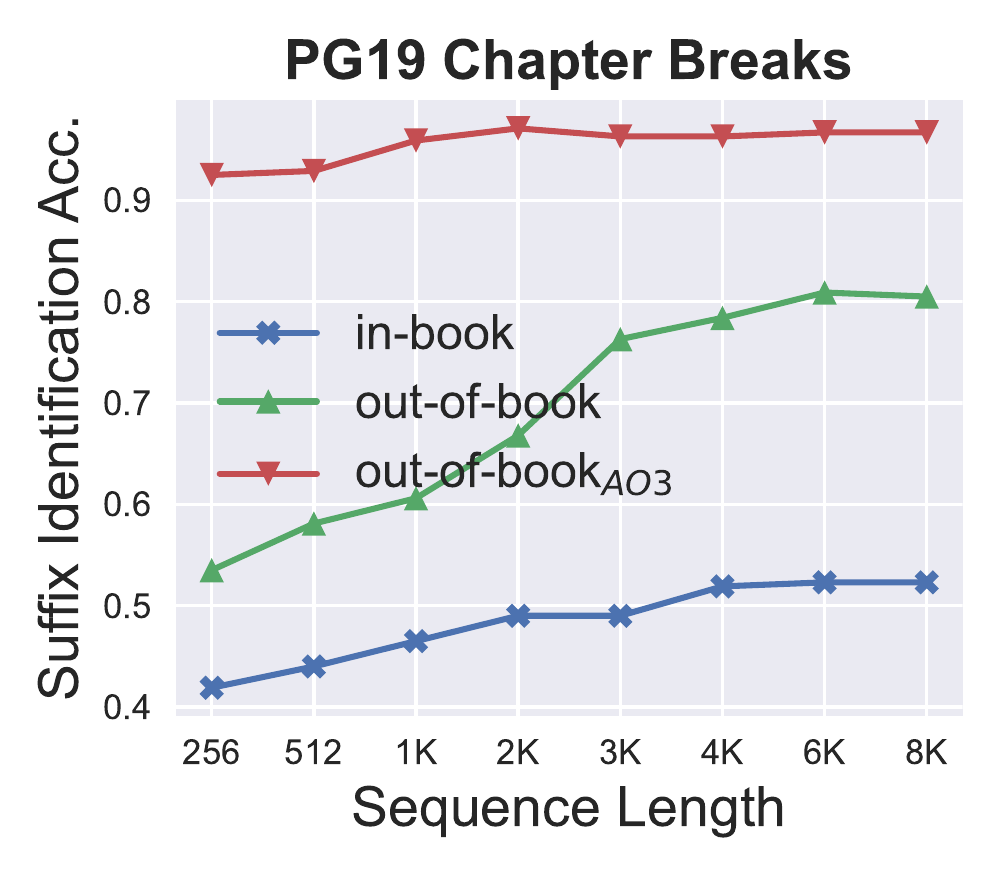}\includegraphics[width=0.215\textwidth]{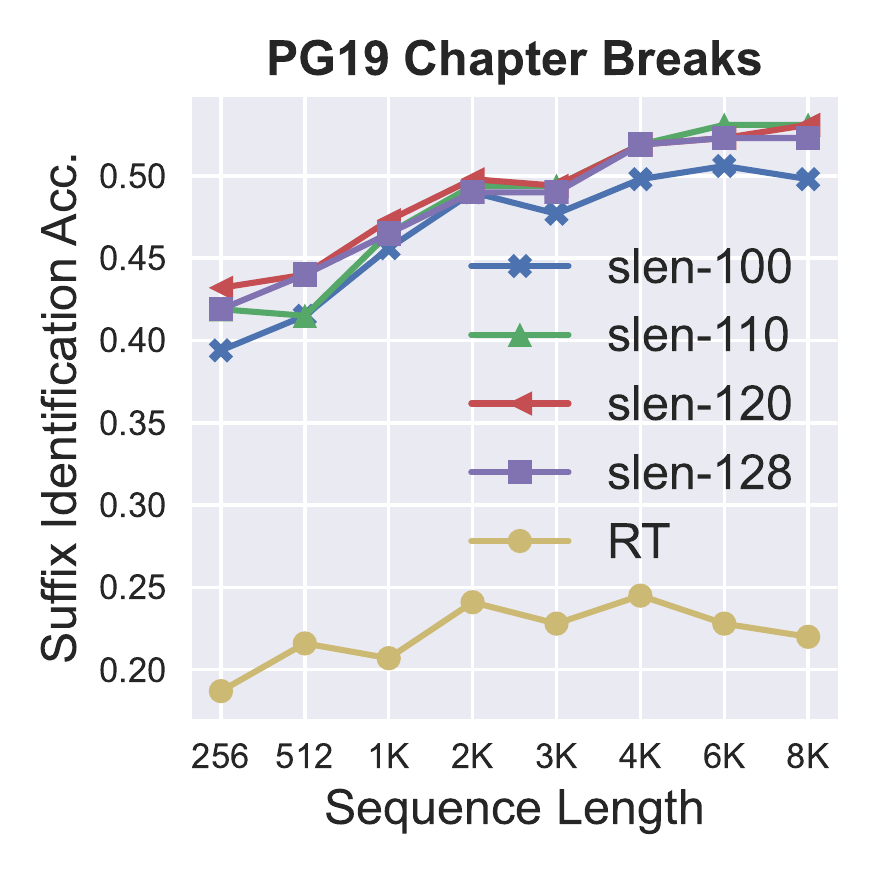}
    \caption{\textbf{Left:} In-book vs. out-of-book. \textbf{Right:} SuffixLM performance when evaluated with different suffix length. The variation in suffix length does not explain the large gap between SuffixLM and token-level LMs.}
    \label{fig:in_vs_out_of_book}
\end{figure}

This section is better read after reading through \S~\ref{sec:experiments}. In this analysis experiment, we show why it is better that the negatives are from the same narrative as the gold suffix. We evaluate our upper bound model SuffixLM on PG-19 set when the negatives are out-of-book suffixes, and plot the suffix identification accuracy in Figure~\ref{fig:in_vs_out_of_book}. When evaluate against out-of-book negatives, this suffix identification task is almost solved by our SuffixLM, especially when the out-of-book examples are from another split in \chbrk. The extremely high accuracy under out-of-book setup suggests the segment representation from different books are easy for SuffixLM to distinguish, thus we adopt a harder setup where the negatives are from the same book. Besides, in-book negatives may contain the same re-occurring named entities or rare words, which require solid understanding of the prefix to differentiate the gold from the distractors.

\section{Various Discourse Relationships} \label{sec:multi-suffix-type}
In addition to chapter breaks, we also evaluate the other two types of discourse boundary examples introduced in Appendix~\ref{sec:data-stats}. As shown in Figure~\ref{fig:all_types}, for all suffix types other than chapter breaks, the evaluated models stop improving as the sequence length grows to more than 2K tokens long. However, there is a significant increasing trend in chapter breaks for SuffixLM. For the rest models, the performance is either flat or not improving. On the AO3 split, the accuracy of SuffixLM improves for $\sim15\%$ as the sequence length increases from 256 to 8K, whereas the improvement of RT is only $\sim1.4\%$. This is in contrast with SuffixLM's $\sim1.5\%$ and RT's $\sim0.3\%$ improvement for the `cause' examples. We draw two conclusions from these observations: (1) the chapter breaks examples form a special case where longer prefix is preferred in order to pick the correct continuation. (2) By comparing the relative improvement, the token-level LMs fall far behind the SuffixLM, which is, besides the absolute performance gap, another evidence that current LRLMs do not effectively leverage long-range context for sequence tasks requiring discourse-level understanding.

\section{Tackle difference in Tokenizers}

As the models we evaluated use different tokenizers, there are small variations in term of suffix length, i.e., the $128$-token suffix may cover different number of words. To understand how the difference in length impacts validity of evaluation, we evaluate SuffixLM with various suffix lengths. Figure~\ref{fig:in_vs_out_of_book} (right) indicates even though there are small variances when the suffixes are of different lengths, the large gap between SuffixLM and Routing Transformer still remains, thus the difference in suffix length does not explain the large performance gap.

\section{Error analysis} \label{sec:err_analysis}

\paragraph{Models struggle with location and event shifts:} Among the 300 examples we annotated in Section~\ref{sec:dataset}, 89 examples were wrongly predicted by all models we have evaluated. By breaking the incorrectly predicted examples into category as presented in Table~\ref{tab:annot_stats}, we find that models tend to make wrong prediction when there is a shift in location or event, and when plots are continuous in timeline.\footnote{ Detailed numbers are included in Appendix~\ref{sec:err_analysis}.}

\begin{table}[ht]
    \centering
    \footnotesize
    \scalebox{0.85}{
    \begin{tabular}{@{}p{0.15\linewidth}p{0.75\linewidth}p{0.1\linewidth}@{}}
\toprule
                      \textbf{Category}       &            \textbf{Definition}                                            & \textbf{Ratio} \\ \midrule
\multirow{3}{*}{Events}      & Previous event ends and new event starts               & 0.74             \\
                             & Previous event continues into next chapter & 0.26             \\ \midrule
\multirow{2}{*}{Actors}      & Change of perspective or character in focus            & 0.43             \\
                             & No change in POV or main character                     & 0.57             \\ \midrule
\multirow{2}{*}{Locations}   & Change of location                                     & 0.64             \\
                             & No change in location                                  & 0.36             \\ \midrule
\multirow{5}{*}{Continuity} & Discontinuous but chronological                        & 0.24             \\
                             & Continuous                                        & 0.62            \\
                             & Analepsis                                              & 0.03              \\
                             & Parallel                                          & 0.11              \\ \bottomrule
\end{tabular}
    }
    \caption{Human annotation on 89 examples sampled from \chbrkao where all models make the wrong prediction. 74\% errors come from the examples where new event starts from the new chapter and 64\% errors from the change of location.}
    \label{tab:err_analysis}
\end{table}

\end{document}